%% file: selfsuplearning_workshop.tex
\documentclass[10pt,twocolumn,letterpaper]{article}

\usepackage{iccv}
\usepackage{times}
\usepackage{epsfig}
\usepackage{graphicx}
\usepackage{amsmath}
\usepackage{booktabs}
\usepackage{amssymb}
\usepackage{arydshln}
\usepackage{mathtools}
\usepackage[numbers,sort]{natbib} 
\usepackage[T1]{fontenc}

\usepackage[pagebackref=true,breaklinks=true,letterpaper=true,colorlinks,bookmarks=false]{hyperref}

 \iccvfinalcopy 

\def\iccvPaperID{31} 

\newcommand{\figref}[1]{Fig.~\ref{#1}}
\newcommand{\secref}[1]{Section~\ref{#1}}
\usepackage{subfigure}

\newcommand{\tabref}[1]{Table~\ref{#1}}

\makeatletter
\def\adl@drawiv#1#2#3{%
        \hskip.5\tabcolsep
        \xleaders#3{#2.5\@tempdimb #1{1}#2.5\@tempdimb}%
                #2\z@ plus1fil minus1fil\relax
        \hskip.5\tabcolsep}
\newcommand{\cdashlinelr}[1]{%
  \noalign{\vskip\aboverulesep
           \global\let\@dashdrawstore\adl@draw
           \global\let\adl@draw\adl@drawiv}
  \cdashline{#1}
  \noalign{\global\let\adl@draw\@dashdrawstore
           \vskip\belowrulesep}}
\makeatother

\DeclarePairedDelimiter\floor{\lfloor}{\rfloor}

\ificcvfinal\fi
\begin{document}

\title{
Self-supervised learning of class embeddings from video\\
}

\author{Olivia Wiles\\
University of Oxford\\
{\tt\small ow@robots.ox.ac.uk}
\and
A. Sophia Koepke\\
University of Oxford\\
{\tt\small koepke@robots.ox.ac.uk}
\and
Andrew Zisserman\\
University of Oxford\\
{\tt\small az@robots.ox.ac.uk}
}

\maketitle

\begin{abstract}
\noindent This work explores how to use self-supervised learning on videos to learn a  class-specific image  embedding that encodes pose and shape information.
At train time, two frames of the same video of an object class (e.g.\ human upper body) are extracted and each encoded to an embedding.
Conditioned on these embeddings, the decoder network is tasked to transform one frame into another.
To successfully perform long range transformations (e.g.\ a wrist lowered in one image
should be mapped to the same wrist raised in another), we introduce a  hierarchical probabilistic network decoder model.
Once trained, the embedding can be used for a variety of downstream tasks and domains.
We demonstrate our  approach quantitatively on three distinct deformable object classes --  {\em human full bodies},
{\em upper bodies}, {\em faces} -- and show experimentally that the
learned embeddings do indeed generalise. 
They achieve state-of-the-art performance in comparison
to other self-supervised methods trained on the same datasets,  and approach the performance of fully supervised methods.

\end{abstract}

\input{introduction_workshop.tex}

\input{related_workshop.tex}

\input{method_workshop.tex}

\input{experiments_workshop.tex}

\input{conclusion_workshop.tex}

\subsection*{Acknowledgements}
This work is supported by the EPSRC programme grant Seebibyte EP/M013774/1: Visual Search for the Era of Big Data.

{\small
\bibliographystyle{ieee}
\bibliography{./shortstrings,./vgg_local,./vgg_other,egbib}
}

\end{document}

%% file: introduction_workshop.tex
\section{Introduction}
How much information is needed to learn a representation of an object class?
Do we require separate representations for different aspects: e.g.\ one representation for 3D, another for pose, another for 2D landmarks?
We investigate how to learn a single representation for a given object class that encodes
multiple properties in a self-supervised manner.
This representation can be used for further downstream tasks and domains with minimal additional effort.

We learn this representation -- which we call an {\em image embedding} -- in a
self-supervised manner from a large collection of videos of that object
class (e.g.\ human upper bodies, or talking heads). The principal assumption is that of {\em temporal  coherence} 
-- that frames of the video contain the object class, but {\em no} additional prior auxiliary information is required.

\begin{figure}
\centering
\includegraphics[width=0.98\linewidth]{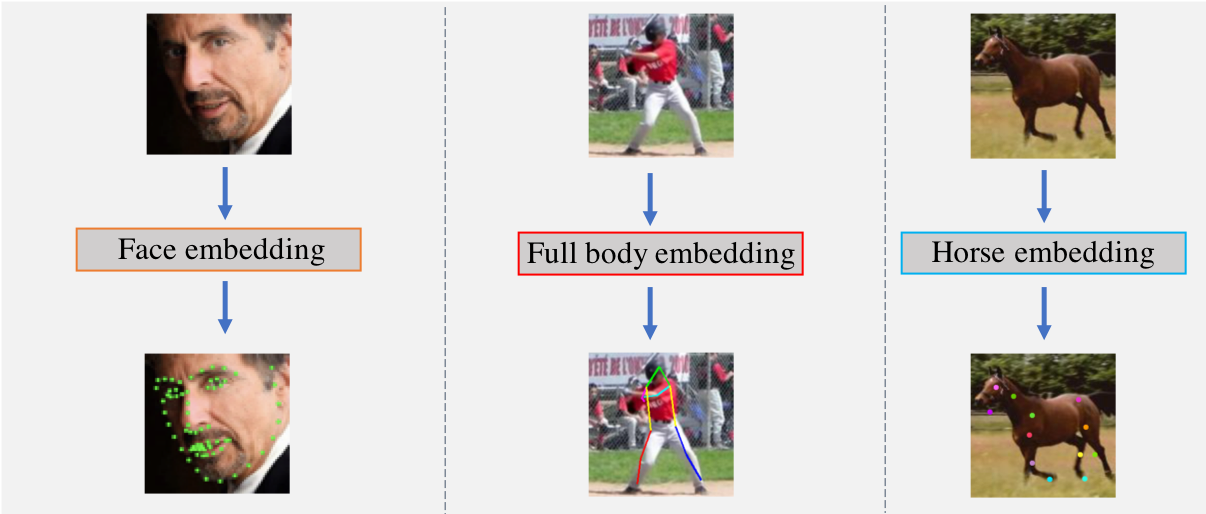}
\caption{The aim of this work is to obtain a class-specific {\em image} embedding
by self-supervised learning on a large
collection of videos. The learned embedding can then be used for a variety of downstream
tasks and datasets.}
\label{fig:teaser}
\end{figure}

In order to learn the image embedding from a video dataset, the following
proxy task is used.  Given two frames from the same video, their image
embeddings  are used to warp one of the frames into the
other.  

We want to model long range dependencies at high resolutions, for 
example large hand movements.
In order to do this, we instantiate the warp probabilistically -- for every pixel in one frame,
we would like to predict the probability that that pixel corresponds to
every other pixel in the other frame.
Doing this naively is computationally prohibitive above a small (e.g.\ $32\times32$) resolution.

As a result, we use a hierarchical approach to perform this operation.
The
model first learns the probabilities at a low resolution,
before refining the probabilities at successive layers while conditioning
on the lower resolution predictions. 
While solving the proxy task at a small resolution may seem trivial,
in fact low resolution images encode important salient information such as spatial layout and context~\cite{torralba2008}.
This approach is inspired by the classical (i.e.\ pre deep learning) multi-resolution methods
employed for optical flow and stereo matching~\cite{Brox09, lucas1981iterative, Adelson84, koch19953}.

The embedding, trained using only pairs of video frames, is
then used for the tasks of predicting landmarks and their visibility on a variety of datasets which may differ substantially
from the initial dataset. 
Our paradigm is useful in applications,
as it requires only one large network per class and one additional
small network per down stream task.

In summary, our contributions are as follows.
\begin{enumerate}
\item A self-supervised class embedding (\secref{sec:selfsuprep}) that can model complex large movements, e.g.\ the movement of arms or hands.

\item 
A hierarchical probabilistic network that allows us to estimate the probability that a given pixel
in a given frame matches each pixel in another frame of the same video for high resolution images. 

\item Two 
additional losses for learning this embedding. The confidence loss (\secref{sec:confidence})
allows the model to express what portions of the target image can be reliably
predicted from the source and what portions cannot. The cyclic loss
(\secref{sec:cyclic}) enforces that the model does not degenerate into
a trivial solution.  

\item We demonstrate that the method learns
a useful representation that can be used for downstream tasks on the same or different domains
for a variety of object classes.
Our method achieves state-of-the-art performance in comparison
to other self-supervised methods trained on the same datasets.
Finally, we show 
qualitative examples of using our approach for a non-human class, that
of horses.
\end{enumerate}

%% file: related_workshop.tex
\section{Related work}
Here, we focus on self-supervised learning from video. 
We also cover class specific modelling, where a model
of the object is extracted using auxiliary information
and then applied to novel
images.

\paragraph{Self-supervised learning on video collections.}
Learning from video \cite{Agrawal15,Jayaraman15,Misra16,Fernando17,Isola15,Mobahi09,Zhou17,Jayaraman16,Pathak17,zhou2015temporal,Vondrick16,Goroshin15,Srivastava15,Taylor10,Wei18} is a powerful paradigm, as unlike with image collections, there
is additional temporal and sequential  information.
The aim of self-supervised learning from video can be to learn to predict future frames~\cite{Vondrick16}, 
or to learn to predict depth~\cite{Zhou17,Garg16,Godard17}. However, we are interested in
learning a set of useful features (e.g.\ frame representations).

One approach is to use the temporal ordering or coherence as a proxy
loss in order to learn the
representation~\cite{Misra16,Isola15,Fernando17,Kim18,Mobahi09,Jayaraman16,zhou2015temporal,Wang16,Wei18}.
Other approaches use egomotion~\cite{Agrawal15,Jayaraman15} in order
to enforce equivariance in feature space \cite{Jayaraman15}. In
contrast, \cite{Jing18} predicts the transformation applied to a
spatio-temporal block.  Instead of enforcing constraints on the
features, one can learn features using a generative task of future
or input frame prediction~\cite{Goroshin15,Srivastava15,Taylor10}.
Another approach is to use colourisation to learn features
and to track objects~\cite{Vondrick18}.

Unlike these works, our focus is to learn a feature representation for
a specific class, which can be used to predict class-specific
attributes.  Most similar to our method is~\cite{Wiles18a} 
which uses video to learn a representation of faces.
However, they do not consider other object classes.

\paragraph{Self-supervised learning of landmarks.}
Instead of using proxy tasks to learn useful features, another line of self-supervised learning is to explicitly
learn a set of landmarks.
This can be done by conditioning image generation on the image landmarks \cite{zhang2018unsupervised,Jakab18}. 
Another approach is to recover object structure by enforcing equivariance to image transformations \cite{Thewlis17b,Thewlis17a}.

%% file: method_workshop.tex
\section{A self-supervised representation}
\label{sec:selfsuprep}
\label{sec:method}

\begin{figure*}
\centering
\includegraphics[width=\linewidth]{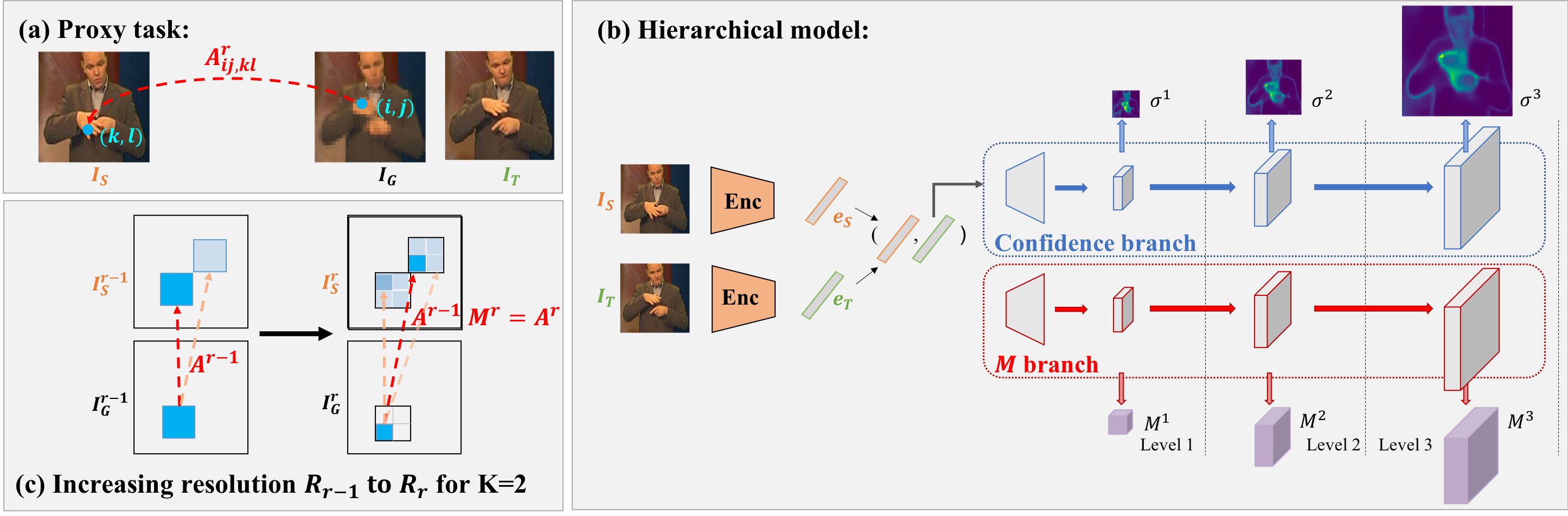}
\caption{An overview of the approach. (a) The proxy task used to train the model. Given a source frame $I_S$ and a target frame $I_T$, 
our model learns a mapping ${\bf A}^r$ to warp the source frame into a generated frame $I_G$. $I_G$ should match $I_T$.
(b) The model in more detail.
The two frames are mapped to embeddings $e_S$, $e_T$. 
Conditioned on these embeddings, the model predicts the warp at an initial resolution $R_1=(32\times32)$ as well as a confidence $\sigma^1$ for each pixel.
These predictions are then refined at successively higher resolutions. 
(c) Illustration of how the predicted ${\bf M}^r$ at each resolution $R_r$ are used to determine the warp ${\bf A}^r$.}
\label{fig:overview}
\end{figure*}

This section introduces our self-supervised model and architecture (\figref{fig:overview}). 
The model is trained for the proxy task of transforming one frame into another frame in a hierarchical manner (\secref{sec:proxy}). 
We allow the model to express uncertainty (\secref{sec:confidence}) and use additional cyclic constraints (\secref{sec:cyclic}) to stop the learned transformation from degenerating.
This gives the final training objective.
We introduce the framework for the case of  human upper bodies, but the same
framework is used for the other classes considered in this paper (full human body, talking faces, horses). 

\subsection{Proxy task to train the network: Modelling the transformation between images}
\label{sec:proxy}
A source frame $I_S$ and a target frame $I_T$ are randomly selected from the same video.
The proxy task to train the model consists of learning how to warp the source frame $I_S$ into the target frame $I_T$. 
Both frames are mapped, using a convolutional encoder with shared weights, to image embeddings
$e_S$ and $e_T$ respectively. 

Conditioned on these embeddings, the model predicts the probability of a pixel in the generated frame $I_G$ matching each pixel in $I_S$.
These probabilities are used to generate the colour of a pixel  by taking the weighted average.
To introduce our notation, let $I_{S_{kl}}$ and $I_{T_{ij}}$ be the colours for pixel locations $(k,l)$ and $(i,j)$ in the source and target frame respectively. 
The network predicts the colour in the generated frame $I_G$ at pixel location $(i,j)$ as a linear combination of pixels in the source frame
\begin{equation}\label{eq:sampling}
I_{G_{ij}} = \sum_{k,l} {\bf A}_{ij,kl} I_{S_{kl}},
\end{equation}
where ${\bf A}_{ij,kl}$ is the probability that a pixel $I_{T_{ij}}$ in the target frame matches a pixel $I_{S_{kl}}$ in the source frame.
We explicitly predict the match similarity ${\bf M}_{ij,kl}$ between a pixel $I_{S_{kl}}$ and $I_{T_{ij}}$ and normalise
the ${\bf M}_{ij,kl}$ to give ${\bf A}_{ij,kl}$ (see Eqs. \eqref{eq:norm1}-\eqref{eq:norm2}).
$I_G$ should match the target frame $I_T$ (\figref{fig:overview}a), which we enforce using a photometric L1 loss
\begin{equation}
\mathcal{L}_{ph}= | I_G - I_T |_1. 
\end{equation}

While using the naive weighted sum works for smaller resolution images, for larger images this becomes computationally prohibitive.
To deal with this problem, we introduce our hierarchical approach
(\figref{fig:overview}b).  Learning 
in a hierarchical manner has been found to be useful in a number of tasks
\cite{wang2018fully,denton2015deep, lai2017deep,
lai2018fast,wang2017growing}.  In our case, the network learns to
determine roughly how to transform points (e.g.\ bigger parts of the
image, like the arms) at a low resolution (${\bf
M}^1$ at level 1).  This transformation is refined progressively at
higher resolutions (${\bf M}^r$ at level $r$).  At these higher levels,
the network can learn to focus on the details
(e.g.\ the placement of the wrists).  This can be regarded as a
form of curriculum learning \cite{bengio2009curriculum} where the
decoder is progressively expanded in levels which increase the
resolution of the generated image.

\paragraph{Probabilistic prediction at a low resolution (Training level $1$).}
At the lowest resolution, $R_1 = (W_1 \times W_1) = (32 \times 32)$, we explicitly predict the probability ${\bf M}^1_{ij,kl}$ that each point $(k,l)$ in the source frame matches each point $(i,j)$ in the target frame.
We then take the weighted average to obtain the probability distribution ${\bf A}^1_{ij,kl}$:
\begin{equation}\label{eq:norm1}
{\bf A}^1_{ij,kl} = \exp({\bf M}^1_{ij,kl}) / \sum_{m,n} \exp({\bf M}^1_{ij,mn}).
\end{equation}
Using the computed probability distribution, we obtain the generated frame (Eq.\ \eqref{eq:sampling}).

\paragraph{Refining the prediction at a higher resolution (Training level $r$.)}

Given the generated frame $I_G^{r-1}$ at resolution $R_{r-1}$, we seek to refine $I_G^{r-1}$ to obtain $I_G^{r}$ at a higher resolution $R_r$.
For a given location $(i,j)$, the highest ${\bf A}^1_{ij,kl}$ give the most likely locations that $(i,j)$ points to in the source frame.
We will use this to limit the locations we consider at the higher resolution (see \figref{fig:overview}c).

In a traditional CNN, as we decode, we would have to keep track of the probabilities for a pixel $(i,j)$ matching every pixel $(k,l)$ in the source frame at that resolution. 
So doubling the resolution of the generated image at each layer requires quadrupling the number of predicted probabilities.
Our insight is that keeping track of all of these probabilities is unnecessary. 
For a given pixel $(i,j)$, we can throw away the unlikely matches at lower resolutions (effectively setting them to $0$) while keeping track of the top $K$ matches. 
Then when we double the resolution at the next layer, we only need to predict $4K$ values (if the width and height of the generated image has doubled, then one pixel at the lower level corresponds to four pixels at the higher level as illustrated in \figref{fig:overview} c)). 
Instead of using these predicted $4K$ values as raw probabilities we use them to re-weight the probabilities predicted at the lower resolution to make the process differentiable.
 This leads to a sparser representation that grows quadratically.

The $M$-branch decoder is used to obtain the $4K$ values ${\bf M}^r$.
These are multiplied by the probabilities at the lower resolution and a softmax normalisation is performed to obtain the final probability distribution ${\bf A}^r$:
\begin{align}
{\bf P}^r_{ij,kl} &= {\bf A}^{r-1}_{\floor{\frac{i}{2}} \floor{\frac{j}{2}}, \floor{\frac{k}{2}} \floor{\frac{l}{2}}}  {\bf M}^r_{ij,kl}\\ 
{\bf A}^r_{ij,kl} &= \exp({\bf P}^r_{ij,kl}) / \sum_{m,n} \exp({\bf P}^r_{ij,mn}).  \label{eq:norm2} 
\end{align}

\noindent{\bf Discussion.}
Our aim is to compute a cost volume that models the probability distribution
of where a pixel in the target frame maps to in the source frame.
\cite{Fischer15} introduced using a cost volume in a deep learning framework for optical flow by 
computing the similarity between features.
This idea has been leveraged in many recent works \cite{Vondrick18,Thewlis16}.
However, naively comparing features at a $W \times W$ resolution requires computing $W^4$ values 
which quickly becomes prohibitively large.
As a result these methods are forced to use a small cost volume or a tiny batch size.

The grid sampler  introduced in~\cite{jaderberg2015spatial} provided
 another way to model the transformation between images by
explicitly learning the warp field.
This was used effectively by \cite{Wiles18a} in order to learn meaningful embeddings for faces.
However, gradients only occur in the local neighbourhood of a point.
As a result if the point needs to travel a large distance between images
and there is no smooth colour transition (as is common in most images), 
then these gradients will be useless and the
model will fail to learn.

Our hierarchical approach gives a way to address the limitation of both approaches.
We can grow the cost volume to image resolutions of the same size as the original image
with minimal overhead.
We additionally do not suffer from the problem of local gradients.
Finally, the hierarchical approach enforces the spatial constraint -- that pixels in a local neighbourhood
move together.

\subsection{Modelling occlusion and background}
\label{sec:confidence}
When modelling the transformation between frames it is possible for part of an object
to become occluded (e.g.\ the hand moving in front of the face) or un-occluded.
Additionally, there may be parts of the scene that are not visible in the previous frame (e.g.\ 
for the signing videos the
background is a video itself and constantly changing).

To allow the model to express uncertainty due to these challenges, we use an
additional decoder which explicitly models the confidence $\sigma^r$ at resolution $R_r$ for the transformation at each location in $I_G$. 
Following \cite{Novotny17b}, we assume that the pixel-wise confidence measure is Laplace distributed and use it to reweigh the photometric loss $\mathcal{L}^r_{ph}$ at each pixel:
\begin{equation}
\mathcal{L}^r_{con} = \sum_{i,j} - \ln \frac{\sqrt{2}}{2 \sigma^r_{ij}} \exp(- \frac{\sqrt{2} |I_{G_{ij}} - I_{T_{ij}} |_1 }{\sigma^r_{ij}}).
\end{equation}

\subsection{Dealing with multiple modes}
\label{sec:cyclic}
One of the degeneracies that can occur when using the probabilistic
approach is a non-injective mapping due to
multiple colour modes (e.g.\ the  three skin regions --
two hands and the head).  For example, a point on the left hand can be
mapped to either hand or the head;
the model is not forced to choose correctly between them.  In
practice, the model cheats and maps all these modes to the one that
moves the least (the head).

The key idea here is to use a cyclic loss~\cite{sundaram2010dense,zhou2016learning}
and normalisation
to enforce uniqueness in order to avoid this problem.
If pixels are transformed from $I_S^1$ to $I_T^1$ and back to $I_S^1$, 
then they should end up at their original location.
If they do not, then it means multiple points in one image are mapped to the same point in another.

The cyclic loss enforces that pixels should return to their original location.
It is formulated as the log likelihood of the 
expectation that a point in the source frame will end up back at the same point at level 1 of the
hierarchical model,

\begin{equation}
\mathcal{L}_{cyc} =  \frac{  \sum_{kl} - \ln (\sum_{ij} (A^1_{kl,ij} A^1_{ij,kl}) )}{W_1 W_1}.
\end{equation}

The loss is minimised when each pixel $(k,l)$ in the source frame maps with probability $1$ to a point in the target frame and that same point in the target maps with probability $1$ to the original point in the source, i.e.\ when ${A}^1_{ij,kl}={A}^1_{kl,ij}=1$.

To enforce uniqueness of the pixel transformation (e.g.\ that not all points in the source frame are mapped to the same point in the target), we perform a normalisation step before applying the cyclic loss.
Points that map to many others in either the source or the target frame are downweighted to give ${A}^1_{ij,kl}$:
\begin{equation}
A^1_{ij,kl} = \min(\frac{{\bf A}^1_{ij,kl}}{\sum_{m,n}{\bf A}^1_{mn,kl}},\frac{{\bf A}^1_{ij,kl}}{\sum_{m,n}{\bf A}^1_{ij,mn}}).
\end{equation}
The matches that still have a high probability are unique in both target and source, as required.

\section{Architecture and training}
All self-supervised models are trained using 3 levels with the lowest resolution $R_1=(32 \times 32)$ which is increased to resolution $R_3=(128 \times 128)$ (as we found additional levels led to marginal improvements). 
They are trained with $K=9$, $\lambda=1$, a learning rate of $0.001$ and the Adam optimizer \cite{Kingma14}. 
When sampling frame pairs from the video, we sample within a distance of $50$ frames from the initial frame for upper body and horses, $20$ frames for full human body and the whole face track for faces. 

\noindent{\bf Architecture.}
We use a convolutional architecture similar to that of \cite{Wiles18a}.
A $256\times256$ image is passed through 8 convolutional layers (interleaved with leaky ReLUs and batch-normalization)
to give a $256D$ embedding.
The confidence and $M$-decoder branches have the same structure but different weights.
The concatenated embeddings are passed through 7 upsampling layers (composed of a ReLU, bilinear upsampler, convolution and batch-norm)
to give a $128\times128$ resolution result.
The intermediary outputs (e.g. ${\bf M}^r, \sigma^r$) are obtained by taking the feature map of resolution $R_r$ and
performing a $5\times5$ convolution to compress the number of channels.

\noindent{\bf Curriculum training strategy.}
The final training objective is the sum of the confidence loss at all layers and the cyclic loss weighted by a hyperparameter $\lambda$,
$\mathcal{L} = \sum_i \mathcal{L}^r_{con} + \lambda \mathcal{L}_{cyc}$.

These losses are trained in a curriculum strategy.
As the predictions of the higher layers depend on those of the lower layers, we train the lower layers to a good local minimum
before training the higher layers.
We start at the lowest resolution $R_1$ and incorporate new layers when the loss plateaus.
The model can first learn a rough estimation of how to transform the source frame into the target before iteratively refining at successively
higher resolutions.

%% file: experiments_workshop.tex
\section{Experiments}
\label{sec:experiments}

We apply the learning framework of \secref{sec:method} to 
three distinct human object classes -- {\em upper bodies},
{\em faces}, and {\em full human bodies} -- to demonstrate its utility by modelling a 
variety of classes with different challenges. In addition to that, we show that our framework is useful for other, non-human object classes by presenting qualitative results for horses.
The question we are seeking to answer here  is whether the embedding that we learn from a 
large set of videos for each object class has encoded useful information about pose and shape of the object. 

\noindent {\bf Downstream learning setup.}
Given an embedding learnt using self-supervision on one of the large video datasets, 
a regressor is trained to map this embedding to the downstream task (e.g.\ landmark prediction). 
This regressor is trained and then evaluated on the given train and test sets of the given dataset. 
For the regressor we consider a linear layer or a multi-layer perceptron containing two 
layers.  While our embedding should learn about pose and expression, there is no reason to expect that the explicit landmarks should be linearly related to the embedding (this is unlike \cite{Jakab18}, which explicitly encode landmarks in their latent representation).
Note that we are {\em not} training our encoder/embedding but {\em only} this regressor.

\noindent {\bf Training datasets.}
The upper body embedding is trained on the Extended BBC Pose dataset \cite{Pfister14a,Charles13} of people signing.
The face embedding is trained on the VoxCeleb2 dataset \cite{Chung18a} consisting of faces of people being interviewed.
The full body embedding is trained on the Penn Action dataset \cite{Zhang13} of people performing sporting actions.
The horse embedding is trained on the horse subset of the TigDog dataset~\cite{delpero16ijcv}.
As our task is not to perform the detection but to learn a representation of the object class,
we use the crops provided by the dataset or, if this is not 
available, a rough crop based on the provided information.

\noindent {\bf Baselines.}
We compare to two baselines.  
The first is using our encoder with random weights; 
this baseline shows how well our self-supervised training improves
over random initialisation.
 The second baseline is \cite{Wiles18a} which
uses a similar proxy task and capacity but a different loss
function/architecture to learn the image embedding and a bilinear sampling for the transformation. They do not use a
hierarchical approach or confidence predictions.
We retrain \cite{Wiles18a} on upper body pose and fully body pose datasets using the authors' code provided
online.

\noindent {\bf Other methods.}
We also report the results of other self-supervised and supervised methods on these datasets.
These approaches vary in terms of how they pre-process their training data and assumptions made about the downstream task.
We give these numbers to benchmark our approach against recent progress but note that these setups are not precisely the same.

\subsection{Predicting landmarks}
\label{sec:landmarks}
We consider the downstream task of predicting landmarks from our learnt embedding.

\noindent {\bf Evaluation metric.}
In order to evaluate the landmarks on upper body and full human body, we use the PCK metric \cite{yang2013articulated}.
This metric reports the percentage of correct keypoints within a normalised distance of the ground truth.
The normalised distance depends on the dataset.
In the case of BBC Pose, we use $d = 6$ pixels as is customary on this dataset.
For FLIC we use a threshold of $0.2 \alpha$ where $\alpha$ is the torso diameter \cite{sapp2013modec}.
For Penn Action we use a threshold of $0.2 \max(s_w, s_h)$ where $s_w, s_h$ are the width and height of the bounding box.
For faces, we report the root mean squared error normalised by the interocular distance.

\subsubsection{Upper body}
\input{./fig_pose.tex}
We use the embedding trained on the BBC Pose dataset to predict upper body landmarks
on the same dataset and on the FLIC dataset \cite{sapp2013modec}. Quantitative results
are discussed below, and qualitative results are shown in \figref{fig:landmarkspose}.

\paragraph{BBC Pose.}
The results on BBC Pose are given in \tabref{tab:bbcpose}.
We first  ablate our approach, demonstrating the utility of predicting confidences, and of using the cyclic loss $\mathcal{L}_{cyc}$.
Each addition improves the average results and the results on the most challenging joint, the wrists.
Using three levels as opposed to one improves performance, demonstrating the utility of the hierarchical approach. 

In comparison to other self-supervised methods, our approach exhibits strong performance.
It performs better than the baseline methods and  \cite{Jakab18}, which was engineered to extract landmarks.
\cite{Wiles18a} fails on this dataset 
due to the problem of local gradients -- the movement between frames (e.g.\ of the hand) during training is too
large, and it degenerates to predicting the identity transformation.
Our approach is also better or competitive with most of the supervised
methods.  Clearly our embedding has indeed learned a semantically
meaningful representation.

\begin{table}[]
\scriptsize
\centering
\caption{Upper body landmark prediction on BBC Pose. Results reported are the PCK for $d < 6$. Higher is better.
 $^\dagger$ denotes training with Extended BBC Pose, else with BBC Pose. The column {\em Loss} specifies the training losses used, $\mathcal{L}_{ph} \text{(ph)}, \mathcal{L}_{cyc}$ (cyc) and $\mathcal{L}^r_{con}$ (con).
 $r$ denotes the level/resolution at which training is stopped. $r=1$ corresponds to a generated image of size $32 \times 32$,
 $r=3$ to a generated image of size $128\times128$.
}
\label{tab:bbcpose}
\begin{tabular}{l l l | c c c c c}\toprule
Method & Loss & Rg. &  Hd & Wrt & Elb & Shldr & Avg \\ \midrule
{\bf Ours} & & & &  \\
r=1$^\dagger$ & ph,cyc,con & lin & 93.7 & 35.8 & 72.3 & 81.6 & 67.7 \\ 
r=1$^\dagger$ & ph,cyc,con  & 2 lr & 94.2 & 51.2 & 78.7 & 82.4 & 74.1 \\ \cdashlinelr{1-8}
r=3 & ph,cyc,con & lin & {\bf 98.0} & 30.7 & 78.9 & 71.3 & 65.6 \\ 
r=3 & ph,cyc,con & 2 lr & 96.5 & 41.0 & 82.4  & 73.2 & 69.9 \\
r=3$^\dagger$ & ph & 2 lr & 94.3 & 54.1 & 79.1 & {83.2} & 75.3 \\
r=3$^\dagger$ & ph,con & 2 lr & 96.0 & 58.3 & {\bf 83.5} & {\bf 83.7} & 78.1 \\
r=3$^\dagger$ & ph,cyc,con & 2 lr & 96.8 & {\bf 62.1} & {82.1} & 82.8 & {\bf 78.7} \\ \midrule
{\bf Self-supervised} & & & &  \\ 
FAb-Net \cite{Wiles18a}$^\dagger$ & & 2 lr & 73.8 & 21.8 & 64.7 & 61.& 52.9 \\
Rand.\ init$^\dagger$ & & 2 lr & 73.2 & 23.2 & 64.5 & 54.7 & 51.1 \\
Jakab {\it et al.} \cite{Jakab18} & & lin & 81.1 & 49.1 & 53.1 & 70.1 & 60.7 \\ \midrule
{\bf Supervised} & & & &  \\ 
\multicolumn{2}{l}{Yang and Ramanan \cite{Yang11}} & & 63.4 & 53.7 & 49.2 & 46.1 & 51.6 \\
\multicolumn{2}{l}{Pfister {\it et al.} \cite{Pfister14a}} & & 74.9 & 53.1 &46.0 & 71.4 & 59.4 \\
\multicolumn{2}{l}{Chen and Yuille \cite{Chen14b}}  & & 65.9 & 47.9 & 66.5 & 76.8 & 64.1 \\
\multicolumn{2}{l}{Charles {\it et al.} \cite{Charles13}}  & & 95.4 & 73.9 & 68.7 & 90.3 & 79.9 \\
\multicolumn{2}{l}{Pfister {\it et al.} \cite{Pfister15a}} & & 98.0 & 88.5 & 77.1 & 93.5 & 88.0 \\ \bottomrule
\end{tabular}

\end{table}

\paragraph{FLIC.}
Given that our approach outperforms the state-of-the-art on the BBC Pose dataset, we consider how well the  embedding
generalises to a new domain, the FLIC dataset, which consists of the upper body of people in film.
The background and people are very different from the BBC Pose dataset.
As can be seen in \tabref{tab:flic}, our approach generalises well to this new domain, achieving high performance.
Again, using three levels as opposed to one improves performance. 

\begin{table}[h]
\scriptsize
\centering
\caption{Upper body landmark prediction at PCK\@0.2 (as defined in \cite{Newell16}) on FLIC using the embedding trained on Extended BBC Pose.
Higher is better. $^\dagger$The entire model is fine-tuned on the FLIC dataset, whereas we regress {\em only} two layers from the 
embedding. }
\label{tab:flic}
\begin{tabular}{ll | c c c c c}\toprule
Method & Rg. &  Hd & Shldr & Elb & Wrt & Avg \\ \midrule
{\bf Ours} & & &  \\
r=1  & 2 lr & 94.2 & 95.7 & 82.5 & 62.6 & 82.3  \\ 
r=3  & 2 lr & 97.2 & 97.1 & 84.8 & 65.2 & 84.5 \\ \midrule
{\bf Self-supervised} & & &  \\ 
Random init & 2 lr & 85.5 & 90.9 & 77.9 & 65.1 & 79.0 \\ 
S\&L \cite{Misra16}$^\dagger$ & & 98.1 & 93.8 & 87.1 & 69.7 & 86.2 \\ \midrule
{\bf Supervised} & & &   \\ 
Newell {\it et al.} \cite{Newell16}  & & -- & -- & 99.0 & 97.0  & --   \\ \bottomrule
\end{tabular}
\vspace{1em}
\end{table}

\subsubsection{Faces}
The second class we consider is faces.
As this model is trained on VoxCeleb2, which has no annotated keypoints, we 
test the learned embedding by predicting landmarks on a variety of other datasets.
This additionally tests the embedding's generalisability.

Our embedding is used to regress landmarks on the AFLW, 300-W, and MAFL datasets and results are reported in 
\tabref{tab:aflw}. 
For AFLW, we report results on the 5-always visible landmarks (AFLW5) as well as for all 21 landmarks (AFLW21).
Qualitative results are shown  in \figref{fig:landmarks}.

\input{./fig.tex}
Our approach performs better than the baseline methods and other methods designed for predicting landmarks when trained 
with similar data.
Our method even performs better than full frameworks trained (self-supervised or supervised) on the given dataset.

\begin{table}[]
\centering
\scriptsize
\caption{Face landmark prediction error on the 300-W and MAFL, AFLW datasets. 
Lower is better.
$^\dagger$ denotes trained on VoxCeleb 1/2, $^\ddagger$ on VoxCeleb 1.
Note that MAFL is a subset of CelebA and models trained on CelebA are fine-tuned on
AFLW when reporting results on this dataset.
Our embedding is never fine-tuned on these datasets; {\em only} the regressor is trained.}
\label{tab:aflw}
\begin{tabular}{ l | c | c | c | c | c}\toprule
{\bf Method} & Regr. &  {\bf 300-W}  & {\bf MAFL}  & {\bf AFLW5} & {\bf AFLW21}  \\ \cmidrule[0.2em](lr{0em}){1-6}
{\textbf{Self-supervised}} & & & & & \\  \cmidrule[0.01em](lr{0em}){1-6}
{\textit{\textbf{Trained on VoxCeleb2}}} & & & & & \\ 
{\textbf{Ours}} & & & & &\\ 
r=3 & lin & 4.93 & 3.21 & 6.73 & {\bf 7.16} \\
r=1 & 2 lr & 5.42 & 3.55 & 7.30 & 7.84 \\
r=3 & 2 lr & {\bf 4.70} & {\bf 2.98} & {\bf 6.64} & {7.28} \\ \cdashlinelr{1-6} 
FAb-Net \cite{Wiles18a}$^\dagger$ & lin & {5.71} & 3.44 & 7.52 & 8.08 \\ 
Jakab {\it et al.} \cite{Jakab18}$^\ddagger$ & lin & -- & 3.63 & 6.75 & -- \\ 
Jakab {\it et al.} \cite{jakab2019learning}$^\ddagger$ & lin & 5.37 & -- & -- & -- \\ \cmidrule[0.01em](lr{0.1em}){1-6} 
{\textit{\textbf{Trained on CelebA}}} & & & & \\
Jakab {\it et al.} \cite{Jakab18} & lin & -- & {\bf 2.54} & {\bf 6.33} & -- \\
Zhang {\it et al.} \cite{zhang2018unsupervised} & lin  & -- & {3.16} & 6.58 & -- \\
Thewlis {\it et al.} \cite{Thewlis17a} & lin & 9.30 & 6.67 & 10.53 & -- \\ 
Thewlis {\it et al.} \cite{Thewlis17b} & lin & 7.97 & 5.83 & 8.80 & -- \\ \cmidrule[0.2em](lr{0em}){1-6}

{\textbf{Supervised}} & & & & \\ 
MTCNN~\cite{zhang2014facial} & & -- & {\bf 5.39} & 6.90 & -- \\ 
TCDCN~\cite{Zhang16} & &  5.54 & -- & 7.65 & -- \\ 
RAR~\cite{XiaoRobust} & &  {\bf 4.94} & -- & 7.23 & -- \\ \bottomrule

\end{tabular}

\end{table} 

\subsubsection{Full body}

Finally, we test our method on full bodies using the Penn Action dataset \cite{Zhang13}.
The person may be seen from the front or back and performing a 
large variety of deformations which results in an extremely challenging dataset.

\begin{figure}[h]
\begin{center}
\includegraphics[width=0.24\linewidth]{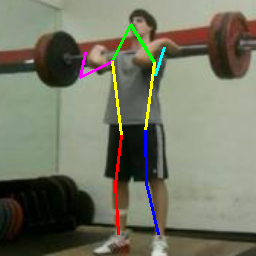}
\includegraphics[width=0.24\linewidth]{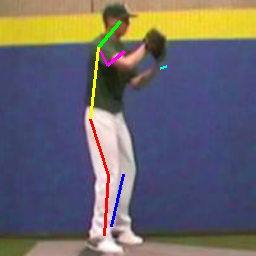}
\includegraphics[width=0.24\linewidth]{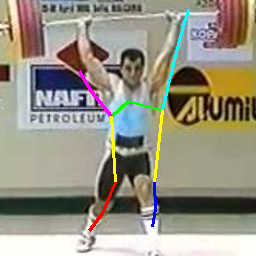}
\includegraphics[width=0.24\linewidth]{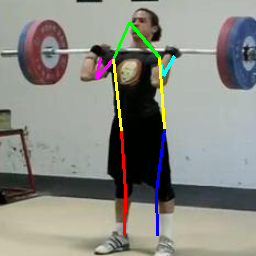}
\end{center}
\caption{Full body 2D landmarks results on the Penn Action dataset.}
\label{fig:fullbodypose}
\end{figure}

We use the learned embedding to regress landmarks.
Quantitative results are reported in \tabref{tab:pennaction} and qualitative results in \figref{fig:fullbodypose}.
We perform better than the baselines, and approach the performance of methods trained
with deep learning on this dataset.
Similarly to upper bodies, \cite{Wiles18a} degenerates to predicting the identity transformation, demonstrating
the effectiveness of our method.

\begin{table}[]
\tiny
\centering
\caption{Full body landmark prediction at PCK\@0.2 (as defined in \cite{song2017thin}) on rhe Penn Action dataset. Higher is better.
}
\label{tab:pennaction}
\begin{tabular}{ll | c c c c c c c c} \toprule
Method & Regr. & Hd & Shldr & Elb & Wrt & Hip & Knee & Ankl & Mean \\ \midrule
{\bf Ours} & & & & & & & & & \\
r=1  & 2 lr & 80.7 & 76.4 & 66.3 & 54.2 & 79.3 & 76.3 & 76.5 & 72.6  \\ 
r=3  & 2 lr & {\bf 83.0} & {\bf 78.8} & {\bf 71.0} & {\bf 58.3} & {\bf 80.9} & {\bf 78.6} & {\bf 76.9} & {\bf 75.1} \\ \midrule
{\bf Self-supervised} & & &  \\ 
FAb-Net \cite{Wiles18a} & 2 lr & 69.3 & 59.1 & 50.2 & 34.0 & 68.8 & 62.2 & 57.5 & 56.4  \\ 
Random init & 2 lr & 70.5 & 60.4 & 50.4  & 35.1  & 70.9 & 63.5  & 53.9  & 56.8 \\  \midrule
{\bf Supervised} & & & &  \\ 
Park and Ramanan \cite{park2011n} & & 62.8 & 52.0 & 32.3 & 23.3 & 53.3 & 50.2 & 43.0 & 45.3 \\ 
Nie {\it et al.} \cite{xiaohan2015joint} & & 64.2 & 55.4 & 33.8 & 24.4 & 56.4 & 54.1 & 48.0 & 48.0 \\
Iqbal {\it et al.} \cite{iqbal2017pose} & & 89.1 & 86.4 & 73.9 & 72.0 & 85.3 & 79.0 & 80.3 & 81.1 \\
Gkioxari {\it et al.} \cite{gkioxari2016chained} & & 95.6 & 93.8 & 90.4 & 90.7 & 91.8 & 90.8 & 91.5 & 91.8 \\
Song {\it et al.} \cite{song2017thin}  & & 97.6 & 96.8 & 95.2 & 95.1 & 97.0 & 96.8 & 96.9 & 96.4  \\  \bottomrule
\end{tabular}
\end{table}

 \subsubsection{Non-human object classes: horses}
 A big advantage of our self-supervised framework is that we can get embeddings for any object class, provided we have video data to train with. To show this, we obtain a horse embedding by training on the horse subset of the TigDog dataset. We train a 2-layer regressor from the embedding to the provided keypoints. Example results can be seen in  \figref{fig:horse}, more results are shown in the supplementary material.
\begin{figure}[h]
\begin{center}
\includegraphics[width=0.95\linewidth]{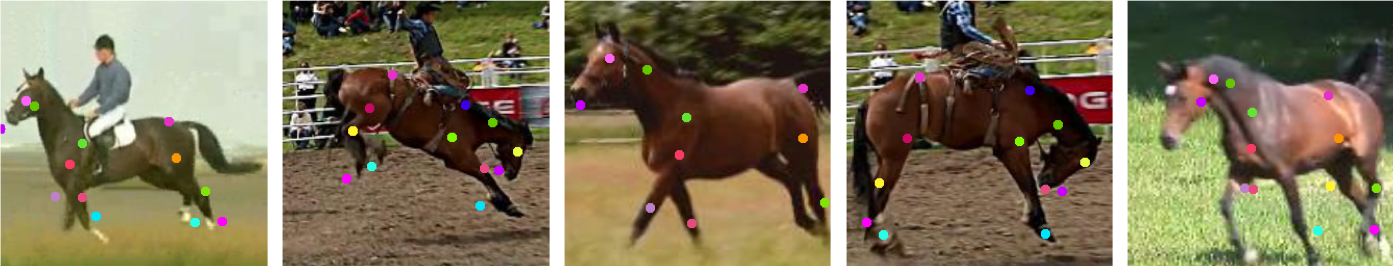}
\end{center}
\caption{2D landmarks results on horses from the TigDog dataset.}
\label{fig:horse}
\end{figure}

\subsection{Predicting visibility}
While we have extensively investigated and demonstrated the high quality of the learned embedding
by using it to regress landmarks, here we investigate whether
the embedding has learned something beyond landmarks.
In particular, we consider whether our embedding can be used to predict whether a
landmark is or is not visible.
Self-supervised methods for detecting landmarks, such as \cite{Jakab18} cannot perform this task, as they
explicitly use the landmarks in their representation.

Both the Penn Action and AFLW datasets have visibility annotations.
We train a 2-layer multi-layer perceptron from the embedding
to predict visibility for each landmark using a binary-cross 
entropy loss.
We compute the area under the curve (AUC) and average over each landmark.
For AFLW, we obtain 89.0 AUC and for Penn Action 77.4 AUC.
A network with random initialisation achieves 63.3 AUC for PennAction and 76.6 for AFLW.
This demonstrates that our method has learned something beyond just 2D positioning.

%% file: fig_pose.tex
\newcommand{\rulesep}{\unskip\ \vrule\ }

\begin{figure*}

\subfigure[{\bf BBC}. Filled dots are GT, empty predictions.]{
\includegraphics[width=0.12\linewidth]{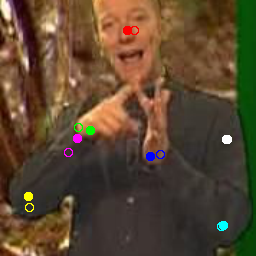}
\includegraphics[width=0.12\linewidth]{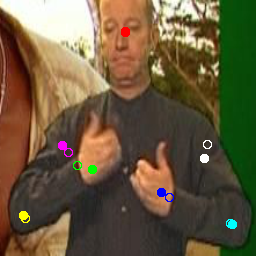}
\includegraphics[width=0.12\linewidth]{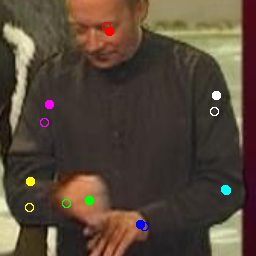}
\includegraphics[width=0.12\linewidth]{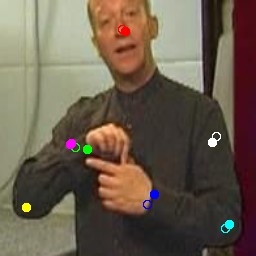}
\includegraphics[width=0.12\linewidth]{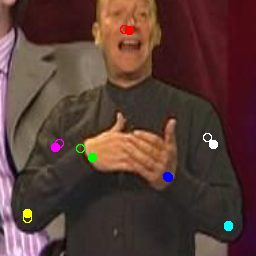}
\includegraphics[width=0.12\linewidth]{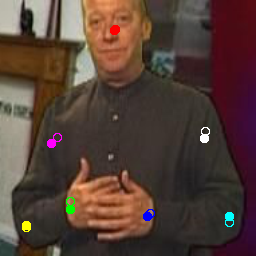}
\includegraphics[width=0.12\linewidth]{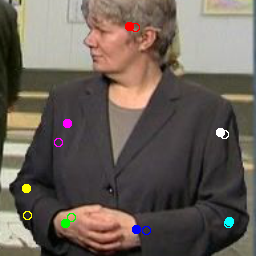}
\includegraphics[width=0.12\linewidth]{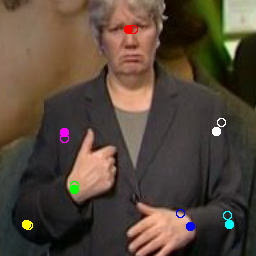}}

\subfigure[{\bf FLIC}. Predicted poses.]{
\includegraphics[width=0.12\linewidth]{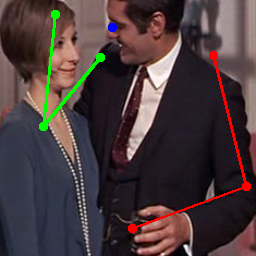}
\includegraphics[width=0.12\linewidth]{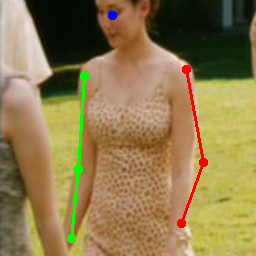}
\includegraphics[width=0.12\linewidth]{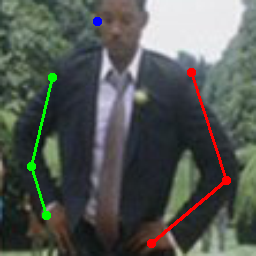}
\includegraphics[width=0.12\linewidth]{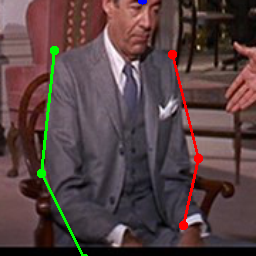}
\includegraphics[width=0.12\linewidth]{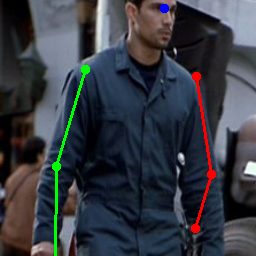}
\includegraphics[width=0.12\linewidth]{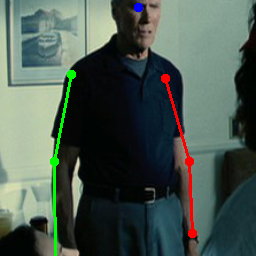}
\includegraphics[width=0.12\linewidth]{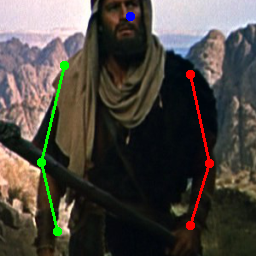}
\includegraphics[width=0.12\linewidth]{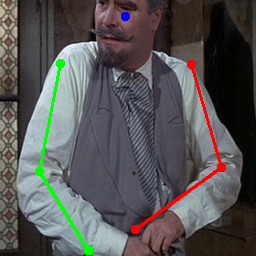}}

\caption{Qualitative results on the upper body pose datasets. More examples are given in the supplementary material.}
\label{fig:landmarkspose}
\end{figure*}

%% file: fig.tex
\begin{figure}[h!]

\subfigure[{\bf MAFL}. Crosses are predictions, dots GT.]{
\includegraphics[width=0.24\linewidth]{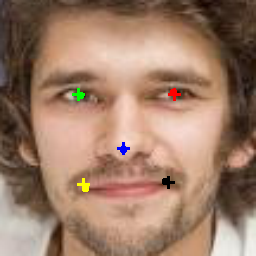}
\includegraphics[width=0.24\linewidth]{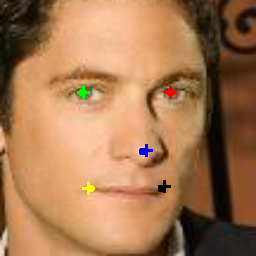}
\includegraphics[width=0.24\linewidth]{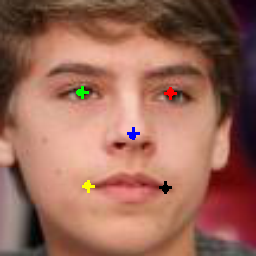}
\includegraphics[width=0.24\linewidth]{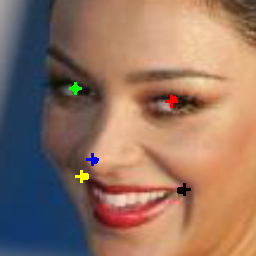}}

\subfigure[{\bf 300W}. Blue is GT, green predictions.]{
\includegraphics[width=0.24\linewidth]{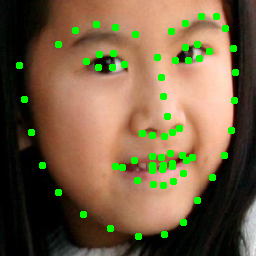}
\includegraphics[width=0.24\linewidth]{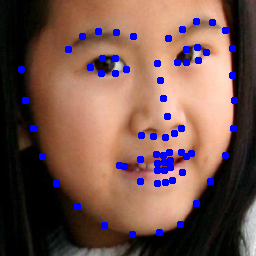}
\includegraphics[width=0.24\linewidth]{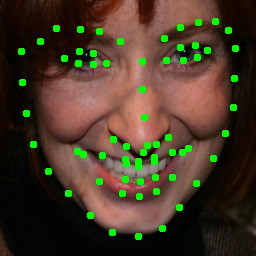}
\includegraphics[width=0.24\linewidth]{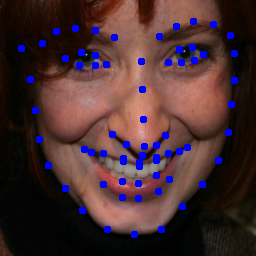}}

\subfigure[{\bf AFLW5}. Crosses are predictions, dots GT.]{
\includegraphics[width=0.24\linewidth]{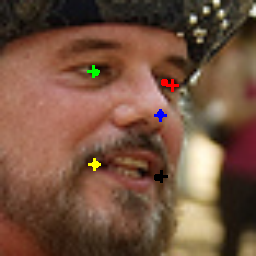}
\includegraphics[width=0.24\linewidth]{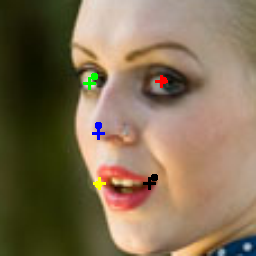}
\includegraphics[width=0.24\linewidth]{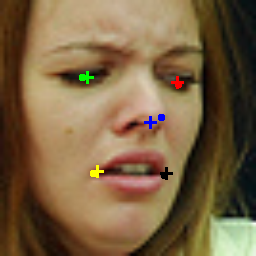}
\includegraphics[width=0.24\linewidth]{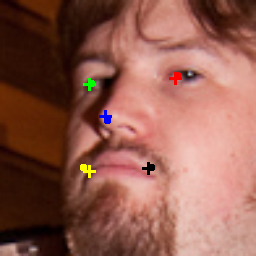}}
\vspace{-0.5em}
\caption{Qualitative results on the face datasets. More examples are given in the supplementary material.}
\label{fig:landmarks}
\end{figure}

%% file: conclusion_workshop.tex
\vspace{-0.5em}

\section{Conclusion}
We have introduced a novel method for 
learning an embedding which encodes high-fidelity 2D landmarks using
self-supervision on video.
Because our method is self-supervised, we can incorporate an unlimited amount of data from 
varied domains to improve the learned
embedding and only use a small set of training data in order to learn the mapping
from the embedding to downstream tasks or domains.
We explore further in the supplementary material how the downstream 
performance varies with the size of this downstream training set. We have demonstrated the method for four distinct 
deformable or articulated classes, but it is equally applicable to rigid classes (e.g.\ cars).

There are many interesting future directions.  The embedding can be learnt for more animal classes 
and used for other downstream tasks.
Also, the embedding could be extended to incorporate the temporal component implicit in the video
in order to summarise multiple frames.